\documentclass[sn-basic]{sn-jnl}

\usepackage{graphicx}%
\usepackage{multirow}%
\usepackage{amsmath,amssymb,amsfonts}%
\usepackage{amsthm}%
\usepackage[title]{appendix}%
\usepackage{xcolor}%
\usepackage{textcomp}%
\usepackage{makecell}%
\usepackage{manyfoot}%
\usepackage{booktabs}%
\usepackage{natbib}
\usepackage{afterpage}
\usepackage{algorithm}%
\usepackage{algorithmicx}%
\usepackage{algpseudocode}%
\usepackage{listings}%

\theoremstyle{thmstyleone}%
\def\rot{\rotatebox}

\theoremstyle{thmstyletwo}%

\theoremstyle{thmstylethree}%

\renewcommand{\orcidlogo}{%
  \includegraphics[width=10pt]{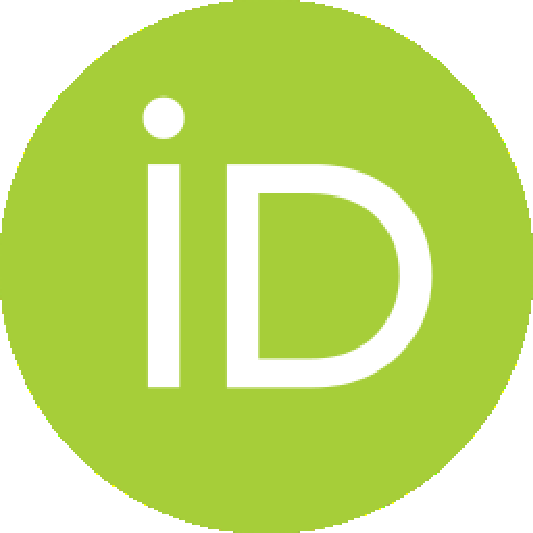}%
}

\begin{document}

\title[Sorting the Babble in Babel]{Sorting the Babble in Babel: Assessing the Performance of Language Detection Algorithms on the OpenAlex Database}


\author*[1,2]{\fnm{Maxime Holmberg} \sur{Sainte-Marie}}\email{mhsm@sdu.dk}

\author[2]{\fnm{Diego}\sur{Kozlowski}}

\author[2,3]{\fnm{Lucía}\sur{Céspedes}}

\author[2,3,4]{\fnm{Vincent} \sur{Larivière}}

\affil*[1]{\orgdiv{Department of Design, Media, and Educational Science}, \orgname{University of Southern Denmark}, \orgaddress{\street{1 Universitetsparken}, \city{Kolding}, \postcode{DK-6000}, \country{Denmark}}}

\affil[2]{\orgdiv{UNESCO Research Chair on Open Science}, \orgname{École de Bibliothéconomie et des Sciences de l'Information (EBSI), Université de Montréal}, \orgaddress{\street{Case Postale 6128, Station Centre-Ville}, \city{Montréal}, \state{Québec}, \country{Canada}, \postcode{H3C 3J7},}}

\affil[3]{\orgdiv{Consortium Érudit}, \orgname{Université de Montréal}, \orgaddress{\street{3744 rue Jean-Brillant, Bureau 6500}, \city{Montréal}, \state{Québec}, \country{Canada}, \postcode{H3T 1P1}}}

\affil[4]{\orgdiv{Observatoire des Science et des Technologies, Centre Interuniversitaire de Recherche sur la Science et la Technologie (CIRST)}, \orgname{Université du Québec à Montréal}, \orgaddress{\street{Case Postale 8888, Succursale Centre-Ville}, \city{Montréal}, \state{Québec}, \country{Canada}, \postcode{H3C 3P8}}}


\abstract{This project aims to optimize the linguistic indexing of the OpenAlex database by comparing the performance of various Python-based language identification procedures on different metadata corpora extracted from a manually-annotated article sample \footnote{OpenAlex used the results presented in this article to inform the language metadata overhaul carried out as part of its recent Walden system launch \citep{Demes_2025, Priem_2025}.}. The precision and recall performance of each algorithm, corpus, and language is first analyzed, followed by an assessment of processing speeds recorded for each algorithm and corpus type. These different performance measures are then simulated at the database level using probabilistic confusion matrices for each algorithm, corpus, and language, as well as a probabilistic modeling of relative article language frequencies for the whole OpenAlex database. Results show that procedure performance strongly depends on the importance given to each of the measures implemented: for contexts where precision is preferred, using the LangID algorithm on the greedy corpus gives the best results; however, for all cases where recall is considered at least slightly more important than precision or as soon as processing times are given any kind of consideration, the procedure that consists in the application of the FastText algorithm on the Titles corpus outperforms all other alternatives. Given the lack of truly multilingual large-scale bibliographic databases, it is hoped that these results help confirm and foster the unparalleled potential of the OpenAlex database for cross-linguistic and comprehensive measurement and evaluation.}

\keywords{OpenAlex, Metadata Quality, Automatic Language Identification, Algorithm Performance Evaluation}



\maketitle

\section{Introduction}\label{sec1}

Automatic Language Identification (henceforth referred to as LI, as in \citep{jauhiainen2024automatic}) generally refers to the task of predicting the natural language or languages present within a given written or speech sample \citep{jauhiainen2019automatic, morris2020automated, jauhiainen2024automatic}. In the specific case of written samples, LI is closely related to the historical practice of deciphering, as both practices make extensive use of statistical techniques based on the compilation and comparison of linguistic patterns and frequencies. But most importantly, both tasks are formally solvable only insofar as their problem space is bounded: not only are the language samples they focus on generated by a grammar whose internal structure is finite, but more importantly, the set of candidate grammars that could have generated these language samples is itself limited \citep{gold1967}. The extraordinary achievements of decipherment, from Egyptian hieroglyphs \citep{champollion1822} to Linear B \citep{ventris1952}, were indeed achievable to the sole extent that decipherers operated within a bounded problem space: ancient scripts contain limited sign inventories, human languages draw from bounded morphological resources, and, crucially, the number of plausible languages an inscription could represent is itself finite, constrained by historical, geographical, and typological factors. In the case of automatic language detection, the problem space is even more constrained: the problem does not consist in retro-engineering long-lost grammars from language samples, but rather in identifying which grammar or grammars generated one or more input strings through comparisons within a closed set of pretrained language models. With a tight-bounded hypothesis space and the possibility of developing and relying on well-modeled distributional models, identifying the language or languages of written samples thus has the potential to become a relatively straightforward task from a statistical and computational perspective, at least in theory.

Attempts to automate this process were developed from the 1970s onward, focusing mainly on resource-rich, frequently-spoken languages \citep{rau1974language, church1985, beesley1988, cavnar1994}. The global expansion of the internet has however significantly broadened the operational scope of LI, as algorithms are now often expected to identify any human language, including low-resource and endangered ones, both in real-time as well as in a wide variety of communication contexts, platforms, and formats \citep{blodgett2022,joshi2020, solorio2014,wang2021}. Both performing and versatile, LI algorithms are now used in various tasks and for many different purposes. As multiple NLP tasks rely on models and tools that are tailored to specific languages, automatic Language Identification (LI) is often used as the first step of processing pipelines, from Machine Translation \citep{goutte2014, zbib2012machine} to web-based, low-recourse language corpus building \citep{jauhiainen2019automatic, jauhiainen2020building, jauhiainen2020experiments, caswell2020language}. However, and as the amount of recent literature on the subject demonstrates, LI is by no means a “solved task” \citep{mcnamee2005language}, since the complexities of real-world data still pose many challenges, particularly with respect to noise and multilingualism \citep{jauhiainen2024automatic}. Perhaps more importantly, several potential areas that could benefit from LI, present new challenges, or offer opportunities for new developments are currently severely under-investigated. The case of multilingual information retrieval is particularly illustrative in this respect. Despite ongoing and significant advances in the field \citep{zhang2023miracl, yang2024language, zhuang2023augmenting, elmahdy2024, goworek2025, ruder2017, jiang2020crosslingual, zhang2022mind}, some obvious and crucial application areas remain surprisingly uninvestigated. The case of identifying the language of content of information records, that is, "the language (or languages) used within the textual components of a resource (i.e. the language that a book is written in)" \cite[1-2]{morris2020automated}, certainly figures among them \citep{knudson2014language, knudson2015automatic-key, chen2014investigation}. For example, a 2018 analysis of the British Library catalog revealed that 4.7 million library records lacked a language code and that only 30\% records from the foundation catalogs were linguistically classified \citep{morris2020automated}. Since very little data of this kind is available, it seems reasonable to assume that such inventories of collection metadata are seldom carried out, which could imply that the lack of information regarding the language of content of the British Library’s documents represents the norm rather than the exception.

A relatively similar situation can be observed in bibliometric databases, although for different reasons. For a long time, automatic language identification simply had little or no bibliometric relevance, as citation indices have traditionally focused on publications written in English \citep{ammon2006language, archambault2005welcome, mongeon2016journal,  demeter2020academic, vera2019web, vanleeuwen2001language, meneghini2007there}. From a language identification standpoint, this skewed language distribution made any attempt at optimizing language metadata rather pointless: one could simply but wrongly assume that all indexed articles are written in English and still get almost all of the cases right; in the same vein, any procedure aiming to adequately identify languages other than English would see their performance gain in that regard be outweighed by the necessarily higher number of English-written articles not identified as such. In a way, this situation can be likened to the one that prevailed within the field of Word Sense Disambiguation. For a long time, algorithms struggled to outperform the main sense heuristic (also called the most frequent sense heuristic), which simply consisted of assigning to each word its most frequent meaning \citep{henrich2015word, abeysiriwardana2024survey, bevilacqua2021recent}. Of course, this situation was mainly due to the skewed nature of polysemy in natural language (the main sense of each word generally accounts for 70\% to 90\% of all use cases) \citep{kilgarriff2004dominant, mccarthy2004finding, navigli2009word, arora2016detecting}), not to the contingent and biased nature of citation indexing practices. However, this more or less explicit partiality of traditional proprietary citation databases toward publications written in English does more than hinder the relevance and use of language identification algorithms. By only compiling references and citations of English-written publications, these databases provide a distorted disciplinary and geographical portrait of the research ecosystem, which misleads every initiative or investigation it informs and affects neglected communities in various ways and degrees \citep{tennant2020web, salatino2021fetichismo, finardi2022linguas, navarro2022rethinking}.

Within the context of this linguistic, cultural, and economic quasi-monopoly, the emergence and growth in popularity of the OpenAlex database feel truly transformational. Driven by its founding values of openness and comprehensiveness, this new citation index already offers the first truly broad and multilingual picture of the global scholarly landscape \citep{priem2022openalex}. Based on an open codebase and on a variety of bibliometric projects, services, databases, and repositories, OpenAlex has been shown to outperform both the Web of Science and Scopus in terms of document and journal frequency \citep{alperin2024analysis, culbert2024reference, van2024oligopoly, jiao2023exclusively}. Of particular relevance here, recent research has shown that OpenAlex also has a higher proportion of articles written in non-English languages than all other bibliometric databases \citep{cespedes2025evaluating, vera2019web}. This stronger allophone content, while making OpenAlex the first citation index likely to benefit from the functionalities offered by automatic language identification, also opens the door for a first thorough evaluation of LI performance in bibliometric context. Such comparative assessment of LI procedures is the focus of the present paper. More precisely, the objective here is to apply different algorithms on various article metadata corpuses extracted from a manually annotated sample of OpenAlex-indexed publications in order to find the algorithm-corpus combination that performs best over a selected set of evaluation measures. The next section will give an overview of the unique challenges posed by Open Alex to current-day LI research, followed by a descriptive and justificatory account of the methodological apparatus designed for the project. The project's results are then presented, analyzed, and discussed in the last two sections.

\section{Research Context}

Both in structure and content, OpenAlex however has the potential to pose some serious challenges to LI research. First of all, not all text attributes that can be found in the article metadata carry language signals, that is, observable cues in text that reflect the grammar, vocabulary, and orthography of of a given language and makes its identification possible \citep{elliott2000language}. In the present project, three different metadata attributes have been considered to carry adequate language signals and thus included in the analysis: title, abstracts, and journal names. However, all three metadata attributes are not always present in all records; some entries only include titles, whereas others include abstracts, journal names, or both. These completeness issues could undoubtedly affect the performance of algorithms whose robustness fails to cope with such metadata irregularities. Also, titles and journal names often contain only a handful of words, thus making them poor carriers of linguistic signal and thus difficult targets for even the best models \citep{jauhiainen2019automatic}. Moreover, research has shown that the severity of this language signal size problem increases in proportion to the number of languages to identify \citep{baldwin2010language}, which makes the task of identifying the language of content of indexed books and periodicals all the more problematic \citep{morris2020automated}, especially in a database as linguistically diverse as OpenAlex \citep{cespedes2025evaluating}

Perhaps even more problematic, OpenAlex is replete with mixed-language content, one of the most challenging issues current-day LI algorithms aim to address. Theoretically-grounded in sociolinguistics, current LI research draws a distinction between two broad types of multilingual context: code-switching, which refers to the alternation between languages across sentence boundaries, and code-mixing, which refers situations in which elements from more than one language are combined within a single sentence, from simple loanwords to linguistically hybrid, yet grammatically coherent crosslinguistic intertwinings \citep{muysken2000bilingual}. Code-mixing is generally considered more complex than its counterpart: cognates (words in different languages that have the same origin and often similar forms), interlingual homographs (words that are spelled the same in two or more languages, but are not etymologically or semantically related), loanwords (words borrowed from another language), and internationalisms (words used globally in many languages) can all undermine the performance of LI algorithms. Such code-mixing can prove especially challenging in the context of scholarly metadata, as it is perhaps more pregnant than other linguistic contexts: proper nouns, technical terminology, words with greek or latin roots (et al., in vitro, quantum,…), and acronyms (CRISPR, AI, LASER, RADAR, SONAR,...), which are often integrated as such across languages, tend to dominate titles, abstracts, and journal names. In addition, modern metadata often relies on publisher-supplied English glosses in parentheses or Latin transliterations of person names, geographic names, or other content not originally written in Latin characters, all which further ambiguate language signals. LI is extremely difficult in such contexts, as language cues point in multiple directions \citep{winata2023decades, goswami2020udldi, morris2020automated}. Moreover, the shorter the number of characters, the more ambiguous or weaker the linguistic cue, and thus the harder to adequately identify the language.

As for code-switching, records frequently include titles followed by a translation in parentheses, or abstracts in two or more languages following each other in sequence. In such cases, boundaries exist between sentences, but metadata schemas tend to not tag them explicitly and include them in a single attribute. A different sort of code-switching can also be found in mixed-language attribute databases like OpenAlex: not only does the same attribute can be in multiple languages across the database, but a single record’s attributes can also be in different languages. This happens for example when a journal’s name or an article title is in a different language than the language of content of the article; Latin and Greek are commonly used for this purpose \citep{morris2020automated}, but modern languages can also be used for that very specific purpose.

For LI research in general, the presence of code-mixing and code-switching implies that the standard “one language per document” assumption, more or less explicitly assumed by most existing research in the field, can no longer be reasonably maintained, which means that systems must shift from monolingual, record-level modeling to finer-grained representations able to handle intra-record multilinguality. This requires rethinking database schemas, information retrieval architectures, and NLP pipelines to support heterogeneous language attributes, dynamic language-specific processing, and multi-label evaluation across all phases of database development and maintenance \citep{jauhiainen2019automatic, Jauhiainen2024PlenarySessions, zhang2023miracl}. OpenAlex does not require such a database overhaul, to the extent that the objective is the same as that of the British Library catalogue project: to identify the main language of content of indexed publications \citep{morris2020automated}. However, this monolingual purpose can still be considerably affected by the mixed linguistic content of a substantial part of OpenAlex’s records. And here, as in other areas that rely on LI, failure to adequately handle multilingualism will unavoidably result in over-representation of dominant languages and lead to serious discoverability problems, since all records whose content language is not properly identified will be omitted from any language-specific user search or query \citep{morris2020automated}. In the context of the predominance of English as the \textit{lingua franca} of Science, both shortcomings will thus tend to reinforce the structural dominance of English in global scholarly productivity and impact \citep{visser2021large, morris2020automated}, henceforth reinforcing global scholarly production inequalities and distorting attempts to measure international research diversity or linguistic coverage of databases \citep{visser2021large}. In light of this, the use of LI algorithms on a multilingual bibliometric database like OpenAlex is more than a simple preprocessing step, but rather a crucial component of responsible bibliometric methodology and epistemic justice.

A final and broader challenge for the application of LI on a large database like OpenAlex pertains to algorithmic feasibility, scalability, incrementability, and implementability, requirements of crucial importance to large-scale databases in general. As OpenAlex contains hundreds of millions of records, the spatial and temporal complexity of LI algorithms, that is, how memory usage and execution time grow as input size increases, is at least as important as how well they manage to identify the language of content of every publication indexed in the database. Moreover and as is often the case with real-world, multilingual systems in general, LI procedures need to be re-run incrementally on the whole database, as its language metadata keeps getting corrected or enriched through the integration of new attributes and records. While increasing the criticality of spatio-temporal constraints, this constant need for updates also makes unusable any procedure based on the retraining of large models.

Considered altogether, the different challenges discussed in this section provide a clear but sobering picture of the situation: OpenAlex record attributes are often short, noisy, inconsistently labeled, with language shifts often occurring within a single attribute, between different attributes of the same record, and even within the same attribute across different records. Given the extensive of bibliometric databases for high-stakes analytics, identifying the language of content of indexed articles is not only an extremely delicate and difficult task, but also one that has the potential to severely distort any picture of the scientific landscape offered by the database, which can in turn profoundly affect science mapping as well as research measurement and evaluation. Ensuring the quality of the linguistic metadata in the OpenAlex database, for the challenges it presents as well as for the issues that arise from it, thus represents a unique and crucial opportunity from both a bibliometric and natural language processing perspective. The next section will describe the method designed here to achieve this goal.

\section{Methods}

Inferring the language of content of an indexed article based on its metadata is no simple procedure, as that language has to be indirectly inferred from the language signals contained in its metadata. Using more metadata attributes than simply titles may help in this task, but the problem of finding the set of attributes that allows for optimal classification is most likely to change depending not only on the algorithm used, but also and perhaps more importantly on the language considered as well as on the performance measure used. In light of these considerations and to cover as much of the whole parameter space as possible, multiple linguistic classification procedures have been designed for this project. A linguistic classification procedure is here defined as the unique combination of an automatic language identification algorithm and a corpus type.

Following various data collection and processing operations, the different metadata corpus types used in this project were generated by first collecting a set of manually and linguistically annotated samples to use as ground truth, then merging each article from the collected sample with different corpora representing all possible combinations of article metadata attributes considered relevant for LI purposes, namely titles, abstracts, and journal names \footnote{Author- and journal-supplied keywords were not indexed in the first version of OpenAlex, most probably for copyright reasons \citep{courtney2024no, carroll2009copyright}. Beyond these legal considerations, however, it could be argued that the inconsistent indexing of keywords, along with their metalinguistic as well as linguistically unnatural and ambiguous nature, all constitute strong arguments against their use for article language identification purposes}.

The different language classification procedures evaluated in this project were then completed by applying each algorithm selected to each and every metadata corpus generated. We then collected performance data for the resulting linguistic classifications by comparing predictions with observations for each annotated article in the sample. Procedure performance is then assessed at the language level first, using a series of evaluation measures commonly used in information retrieval, then at the database level, by simulating the overall performance of each algorithm under different indicator weighing regimes using Bayesian conjugate inferences over probabilistic confusion matrices. These different steps are detailed further in the following subsections.

\subsection{Sample Collection}

The ground truth for this study is based on the annotation work performed by \citet{cespedes2025evaluating} in their assessment of the quality of language metadata in OpenAlex. For that paper, a multilingual sample of indexed articles was manually annotated following a two-step process. First, two samples of indexed articles were randomly drawn from the database, a first one containing 50 articles for each of the 55 languages present in the database (the resulting sample included only $2701$ articles, as some languages had less than 50 indexed articles) and a second containing 300 documents for each of the 11 most frequent languages in the database. Then, each article sampled was manually annotated by not only inspecting its title and abstract, but also by accessing the full text of the corresponding document, identify its language, and then determine if that paper indeed corresponds to a scientific paper; entries whose documents couldn't be accessed were ignored (see \citep{cespedes2025evaluating} for more information about these samples).

In addition to these two samples and in order to account for linguistically-unclassified articles indexed in OpenAlex, a sample of 300 such articles was collected for the purposes of the current project. Following extraction of this third subset, whose size relative to the previous two samples is proportional to the relative frequency of linguistically-unclassified articles in OpenAlex as of July 25th 2024, all articles thus sampled were annotated based on the same procedure as the one used for the two previous samples.

\subsubsection{Corpus Building}

\begin{figure}[ht!]
\centering
\includegraphics[width=\textwidth]{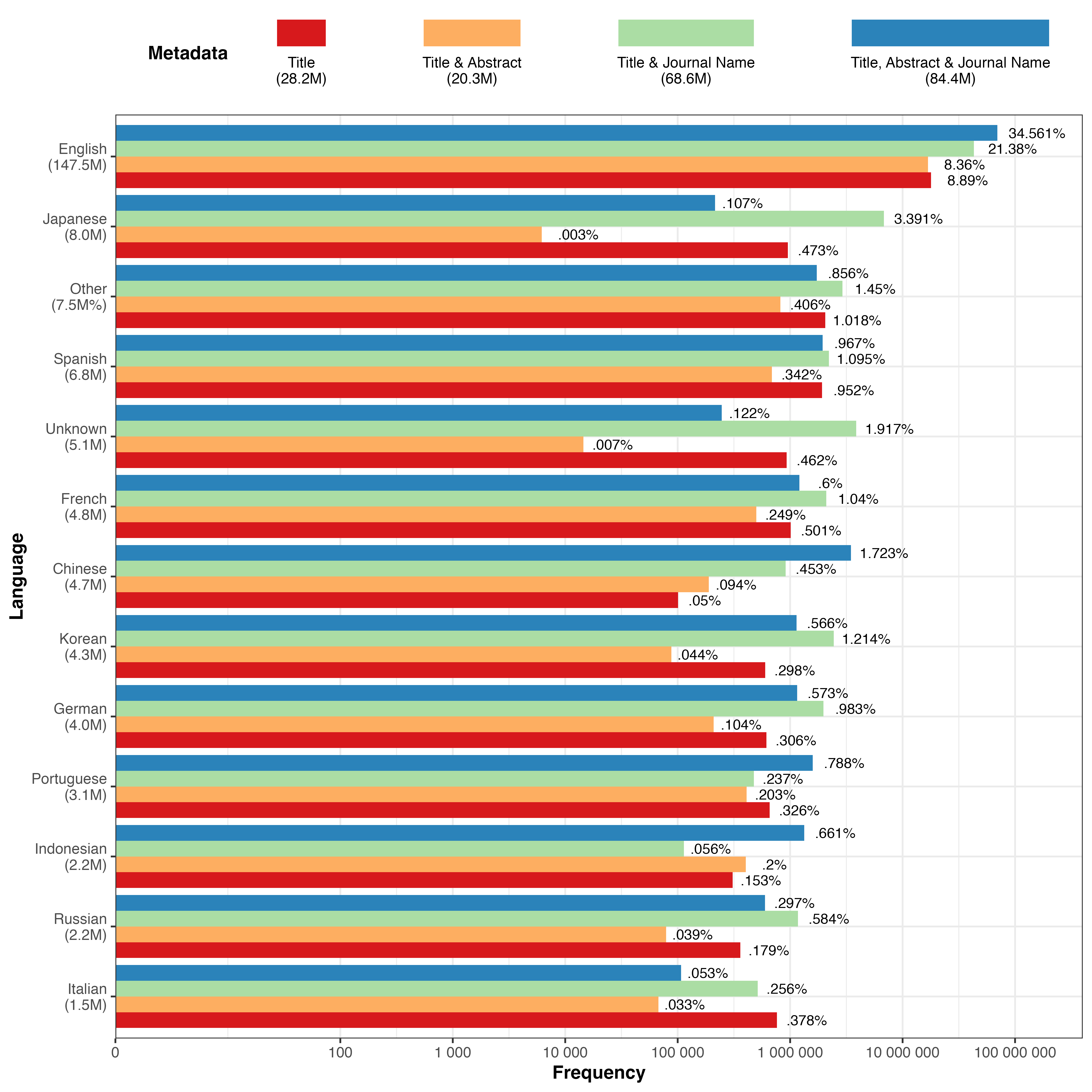}
\caption{Percentage of Articles included in the OpenAlex database, grouped by by Language and Metadata Availability}
\label{fig:langfreq}
\end{figure}

As can be expected of a bibliographic database of such scale, OpenAlex contains articles from a variety of sources, from publishers and various scientific or professional organizations to conferences and self-archiving repositories. As a result, the type and quantity of metadata associated with articles indexed in the database can vary enormously: with regard to textual metadata in particular, some entries only include titles, whereas others include abstracts, journal names, or both. Figure \ref{fig:langfreq} shows the absolute and relative frequency distributions of all articles indexed in OpenAlex as of December 25th, 2025, grouped by language and metadata configuration type. As regards to languages, the 11 most frequent languages in the database are directly represented in the figure, along with two additional categories, which, respectively, include all articles classified in less frequent languages (other) and articles currently not classified linguistically (NA). Regarding configuration types, four different categories are distinguished: articles with titles, articles with titles and abstracts, articles with titles and journal names, and articles with titles, abstracts, and journal names. Absolute frequencies for languages and category types are respectively shown in millions (M) below the different y-axis and legend labels, while relative proportions for each metadata configuration type of each linguistic category are shown in percentages at the end of each corresponding horizontal bar. Finally, the x-axis was scaled algorithmically in order to better compare proportions and frequencies within and across language categories. 

As this figure shows, most of the articles in the database are written in English, as previously reported in \cite{cespedes2025evaluating}, and a majority of these English articles include both abstracts and journal names as metadata. Asian languages form the second most frequent linguistic group, followed by Romance languages. As for the different metadata attributes considered here, no clear cross-linguistic pattern can be found: while all articles have titles as attributes, metadata availability as regards to abstracts or journal names varies a lot across languages. Articles that include both their abstract and the name of the publishing journal represent a minority of indexed articles: besides English, only Chinese and Indonesian have a majority of indexed articles that include both articles and journal names. As for the remaining languages, most article entries only have journal names alongside titles, and a majority of Italian articles do not have any relevant information other than titles.

For the purposes of the present project, this metadata variability is far from trivial. First of all, the performance of the algorithms used here should be expected to strongly correlate with input text length: the more text is given as input, the more accurate the predictions can be expected to be, regardless of the algorithm considered. Additionally, some LI algorithms might be particularly sensitive to specific attribute features; given the constantly growing and changing nature of the OpenAlex database, algorithm performance at the database level at a given phase of its development could very well depend on the saliency and pervasiveness of these antipatterns within the metadata configuration at any given time.

In light of these considerations and to find the language classification procedure whose predictive power is most robust to such changes, different corpora were built for the sampled articles based on the metadata combination types used in figure \ref{fig:langfreq}.

\begin{table}[ht!]
\centering
\begin{tabular}{r|p{5cm}}
\textbf{Corpus Type} & \textbf{Procedure}\\
\hline & \\[-0.15cm]
Titles (T) & For each article, get the title.\\[.15cm]
Titles \& Abstracts (A) & For each article, get its title; if available, get its abstract.\\[.15cm]
Titles \& Journal Names (J) & For each article, get its title; if available, get the name of the journal.\\[.15cm]
Greedy (G) & For each article, get its title; if available, get its abstract and the name of the journal.\\[.15cm]
\hline
\end{tabular}
\caption{Description of the corpus types}
\label{table:corpus}
\end{table}

Each of these corpora is built so as to include at least the title of every article in the annotated sample. This ensures that every procedure creates a complete corpus by assigning one document to each article sampled, which allows LI algorithms to exhaustively partition the sample into the selected language categories. Finally, in order to syntactically separate document attributes in each corpus, text attributes not ending with periods were punctuated accordingly.

\subsection{Algorithms}

All LI algorithms whose performance is assessed here have been selected based on two criteria: Python implementability and completeness. In the first case, only off-the-shelf, python-based, LI systems that do not require any sort of training or parametrization by the user are considered here. The main reason for this is that since OpenAlex is Python-based, evaluating classification procedures implemented in other languages would be unjustified from an implementational perspective, as it would only raise unnecessary feasability issues. This first criterion is not problematic in any case, since most LI algorithms are implemented in Python or can be used in Python via packages specifically developed to that end. Another reason for the use of Python-based algorithms is that the OpenAlex back-end is itself Python-Powered; Large Language Models were also excluded for this reason as well for their inherent scalability limitations, which prevent their use on a large-scale database like OpenAlex. As for the second criterion, an algorithm is said to be complete if it terminates with a solution when one exists \citep{pearl1984heuristics}; in the case of the present project, any algorithm said to be complete is thus one that is able to return a linguistic prediction for each document given as input. Although this criterion does not seem too restrictive at first sight, some algorithms have a relatively low language coverage and have not been designed or trained to detect all the languages present in the OpenAlex database. For this and other reasons, any algorithm that was unable to return a prediction for every document submitted was excluded from subsequent analyses, regardless of its overall performance.

Based on these restrictions, the different LI algorithms selected and evaluated for this are shown and described in Table \ref{table:algo}.

\begin{table}[ht!]
\centering
\begin{tabular}{r|p{9cm}}
\textbf{Algorithm} & \textbf{Description}\\
\hline & \\[-0.15cm]
CLD2 & Second installment of the Compact Language Detector developed by Google for its Chrome browser, CLD2 is built on a Naïve Bayesian classifier trained on a manually constructed corpus of chosen web pages for each language \citep{sites2025compact}.\\[.15cm]
CLD3 & Third installment of the Compact Language Detector designed for Chrome. It is a multi-layered neural network built on word embeddings based on character unigram, bigram, and trigram proportions and able to detect 107 languages \citep{salcianu2025}.\\[.15cm]
DetectLanguage & Partly based on Google's CLD2 Cloud translation API, this LI web service allows its users to detect the language of a text either through a POST request or its API \citep{detectlanguage2025}.\\[.15cm]
FastText & Created by Facebook's AI Research lab for efficient learning of word representations and sentence classification, fastText's language identification model is trained on data from Wikipedia, Tatoeba, and SETimes \citep{joulin2016bag, joulin2016fasttext,bojanowski2016enriching}. Available in both uncompressed and compressed versions, only the uncompressed and more performant version is used here.\\[.15cm]
FastSpell & FastSpell complements FastText with spell-checking functionalities from Hunspell. This algorithm detects languages by first trying to determine the language of a sentence with FastText, then makes extra checks with Hunspell to determine the language more precisely \citep{banon2024fastspell}.\\[.15cm]
Langdetect & A Java-based language-detection library, Langdetect is built on a Naive Bayes classifier, trained on a mix of n-gram-based language-independent character corpora extracted from Wikipedia abstracts \citep{nakatani2010langdetect}\\[.15cm]
LangID & This algorithm uses a Naive Bayes classifier trained on relative frequencies of character n-grams extracted from JRCAcquis, ClueWeb 09, Wikipedia, Reuters RCV2, and Debian i18n \citep{lui2011cross, heafield2015language, lui2012langid}\\[.15cm]
\hline
\end{tabular}
\caption{Description of the different LI algorithms}
\label{table:algo}
\end{table}

For every algorithm assessed here, different linguistic classification procedures were designed by applying it on each metadata corpus generated, resulting in a total of $4 * 7 = 28$ different sample partitions.

\subsection{Performance Assessment}

To assess the predictive power of each classification proceudre as regards to article language, a confusion matrix was built for each of the 12 different linguistic categories (the 11 most frequent languages in the database (those identified in figure \ref{fig:langfreq}, along with a 12th category including all other languages) by determining, for each language sampled, if the language detected by the algorithm and the one observed through manual annotation are the same. In the case of English for example, this procedure consists in determining, for each annotated article, if the metadata language detected by the algorithm for the corresponding article is in English and if manual checking has determined (observed) that the article is indeed written in English. Following this and in accordance with the procedure used in \citet{cespedes2025evaluating}, true positive counts, false positive counts, and false negative counts were compiled and grouped by by classification procedure and language. Then, in order to assess the performance of each LI procedure for each different language category, three different but complementary performance measures were implemented and used.

The first measure, precision, assesses the correctness of an algorithm's predictions. In the present context, the precision of an algorithm for a specific language corresponds to the proportion of articles predicted to be written in that language that are indeed written in that language. In other words, if we were to ask an algorithm for all articles written in German, that algorithm's prediction score for German corresponds to the proportion of articles it identifies as being written in German that are indeed in that language.

Often used as a complement to precision, recall aims to assess how exhaustive an algorithm's predictions are. Here, the recall score of an algorithm refers to the proportion of articles written in a given language that are identified as such by the algorithm. To take the previous example, if we were to ask an algorithm for all articles written in German, the recall score of that algorithm for German corresponds to the proportion of all articles written in German that are returned by the algorithm. Both precision and recall have true positives at the numerator, that is, the number of accurate predictions returned by the algorithm; their distinction lies at the level of the denominator, which respectively corresponds to the total number of positive predictions (true and false positives) and relevant cases (true positives and false negatives).

While precision and recall are the usual go-to indicators in matters of algorithmic performance, processing speed is often what tips the balance at the implementation phase, especially in industrial settings. In the case of the present study, how much time it takes for an algorithm to predict the language of every document of every corpus is of paramount importance, as a difference of a few seconds at the level of the collected sample can easily translate into hours, days or even weeks at the database level. To properly assess procedure scalability, the processing speed of each algorithm was collected by timing how long it takes to process each corpus in its entirety. All collected times were then rescaled to the $[0, 1]$ interval through normalization, which was done by multiplying the time recorded by each procedure on each corpus by the reciprocal of the quickest time obtained by any procedure on any corpus.

\subsection{Simulation}

While the procedures described above provide interesting information on the performance of the various algorithms and allow them to be broken down by language and type of procedure, these descriptions can however be misleading when it comes to assessing the performance of the various procedures at the scale of the entire database as they do not take into account robustness to uncertainty as well as the impact of data imbalance on algorithm performance evaluation. To address these crucial dimensions of algorithm performance, a simulation based on various probabilistic models was carried out and its results presented over various configurations of precision, recall, and processing speed weighing.

\subsubsection{Robustness Assessment}

As results presented in \citet{cespedes2025evaluating} show, the quality of OpenAlex's language metadata is very good; however, it is not perfect, which means that the language frequency distributions returned by the database necessarily differ from the actual ones to an unknown extent. Also, OpenAlex is an incessantly growing and changing database, as new articles are indexed and metadata of already indexed articles gets updated. The current state of the database in terms of language proportions and types of metadata is therefore not only uncertain at the present time, but also bound to change in the more or less short term. In that perspective, algorithmic robustness, that is, the ability of an algorithm to maintain "some desired system characteristics despite fluctuations in the behavior of its component parts or its environment" \citep[p.2539]{carlson2002complexity}, is bound to have a crucial impact of the quality of any linguistic classification of the database. In other words, an evaluation of linguistic classification procedures for the OpenAlex database has to take into account the ability of each procedure to that maintain good overall performance despite uncertainties in and changes to the current state of the database.

In order to best quantify database uncertainty and algorithm robustness, probabilistic confusion matrices were generated based on the sample predictions collected from the different LI procedures. Several studies have already demonstrated the usefulness of Bayesian uncertainty modeling in the context of algorithm performance evaluation \citep{goutte2005probabilistic,brodersen2010balanced,benavoli2017time,caelen2017bayesian,totsch2021classifier,chlaily2023measures}. The probabilistic approach adopted here will consist in modeling uncertainty by treating precision, recall as well as the relative proportion of languages and metadata types in the database as random variables modeled using different conjugate distribution models. In conformity with previous probabilistic modelisations of confusion matrices and submatrices \citep{brodersen2010balanced,totsch2021classifier}, precision and recall probabilities for each algorithm, language, and metadata configuration, were modeled using posterior beta-binomial distributions with uniform priors for the confusion submatrices of each distinct database subgroup $k$; true positive counts were used as alpha parameters for both measures, while false positives and false negatives were used as beta parameters for precision and recall distributions respectively.

\begin{align}
Precision &\sim Beta(TruePositives+1,FalsePositives+1)\\
Recall &\sim Beta(TruePositives+1,FalseNegatives+1)
\end{align}

To model the proportions between the different language and metadata configuration groups within the database, a posterior Dirichlet-multinomial distribution with uniform prior was used, which generalizes the beta-binomial model with Laplace priors to multiple classes or categories, in this case the different database subgroups shown in the mosaic figure above. Proportion probabilities were thus obtained by adding the absolute article frequencies $n_{1...k}$ of each $k$ database subgroup as alpha parameters of the underlying beta distributions and incrementing them by 1 as in the case of precision and recall.

\begin{align}
LanguageSubgroupProportion &\sim Dirichlet((n_1+1)...(n_k+1))
\end{align}

Using these different conjugate distributional models, overall performance scores for the OpenAlex database were estimated through the following simulation procedure: for each classification procedure, precision and recall scores are obtained by drawing a random number from a mixture of beta-binomial distributions build on the classification outcomes for each language category, mixture whose component weights are obtained through a single draw of the language proportion model. This random sampling process is repeated $100\,000$ times, and scores for each measure and procedure are compiled into a single simulation dataset.

\subsubsection{Data Imbalance}

As is often the case in large databases, OpenAlex suffers from severe distribution skews, many of which are caused by the preponderance of English-written articles, which allegedly account for $75.1$\% of all database entries (imbalance ratio of 3:1). Such an asymmetry, apart from the question of its representativeness of the real linguistic distribution of scholarly communication on a global scale, poses a serious problem with regard to the evaluation of the linguistic classification procedures implemented for this project. As is the case with imbalanced sets in general, the general objective is to improve recall without hurting precision; however, the number of false negatives for the minority classes can hardly be reduced without at the same time increasing their false positive count \citep{he2013imbalanced}. Moveover, standard classification algorithms are often biased towards the majority English class, leading to a higher misclassification rate for instances belonging to minority classes, in the present case all other languages \citep{lopez2013insight, liu2006boosting,huang2006intelligent}. In the specific context of the OpenAlex database and as pointed out in \citet{cespedes2025evaluating}, classifying article language in the OpenAlex database thus involves a trade-off between precision performance for English and retrieval performance for all other languages: a lower precision score for English will result in a better performance overall across all metrics, at the expense however of an underestimation of all other languages. This is all the more troublesome as the minority classes, i.e. all non-English languages in the present project, are usually those that are the most crucial from a classification perspective \citep{lopez2013insight}. In the context of the current project, it could also be argued that these minority classes are actually the most interesting ones, as they are what motivates the linguistic classification in the first place.

These imbalanced-related issues cannot be addressed directly here, as the procedures evaluated here are both built on algorithms and applied to samples that were respectively trained and collected beforehand. However, it is possible to get a comparative understanding of the impact of database-level trade-offs between precision and recall for any procedure through weighted harmonic aggregation of the scores it obtained throughout the simulation. To do this and in order to also take into account processing times for each procedure, the following generalization of the $F_{\beta}$ score formula is used here, formula which adds a $\gamma$ parameter to control the relative weight of processing speed relative to precision and recall:

\begin{align}
F_{\beta,\gamma} &= (1 + \beta^2 + \gamma^2) \cdot \frac{precision \cdot recall \cdot speed}{(\beta^2 \cdot precision) + recall + (\gamma^2 \cdot speed)}
\end{align}

Regarding $\beta$ values, $\beta > 1$ and $\beta < 1$ allow $F_{\beta,\gamma}$ to respectively confer more or less importance to recall in relation to precision, with lower and upper bounds $\beta = 0.5$ and $\beta = 2$ respectively conferring to recall half and twice the importance of precision. As for $\gamma$,  $\gamma = 0$ excludes speed from $F_{\beta,\gamma}$, $\gamma = 1$ confers equal weight to both precision and speed, while lower and upper bounds $\gamma = 0.5$ and $\gamma = 2$ makes speed half and twice as important as precision, in that order. Given this, $\gamma = 0$ makes $F_{\beta,\gamma}$ formally equivalent to $F_\beta$, $\beta = 1$ and $\gamma = 1$ returns the standard $F_1$ score, and setting $\beta = 1$ and $\gamma = 1$ makes $F_{\beta,\gamma}$ return the harmonic mean of precision, recall and speed. \citep{sasaki2007truth}. 

Overall simulated performance for each procedure is then obtained by calculating $F_\beta,\gamma$ scores obtained by each procedure at each simulation step for a set of various $\beta$ values ranging from $0.5$ (precision is twice as important as recall) to $2$ (recall is twice as important as precision) as well as 4 different values of $gamma$: 0 (speed is ignored altogether), 0.5 (speed is twice as important as recall), 1 (recall and speed are equally important), 2 (recall is twice as important as speed). Finally, \textit{maximum a posteriori} estimates for each procedure and every $F_{\beta,\gamma}$  weighing configuration are obtained by extracting the mode of each corresponding posterior density distribution.

\section{Results and Analyses}

In this section, the predictive power of the different linguistic classification procedures implemented is first assessed at the sample level: precision and recall scores for the different languages investigated are presented and analyzed, followed by a comparison of the processing speeds recorded by each algorithm over the different corpus types. Finally, the highest-performing procedures for each weighing regime are presented and compared.

\subsection{Performance by Language}

The various scatter plots included in Figure \ref{fig:perf} show performance scores obtained for each language and procedure. In order to simply both visualization and interpretation, only best-performing procedures in either precision or recall were considered for each algorithm. In case of equal score between two or more procedures using the same algorithm, only the best-performing procedure as regards to the other measure was kept (i.e. best recall score for procedures with equal precision score or best precision score for procedures of equal recall score). Additionally, in case of equal precision and recall scores between two or more procedures using the same algorithm, only the one based on the shorter, simpler, and faster-processable corpus was kept (with Titles < Titles \& Journal Names < Titles \& Abstracts < Greedy). Finally, in cases where a procedure from a given algorithm and language obtains the best score for both measures, no other procedures for that algorithm and languages are represented. Precision and recall scores for all procedures implemented for this study can be found in appendix \ref{appendix:perf}.

In each subplot, procedure performance is represented by points whose X and Y coordinates respectively refer to the precision and recall scores of each corresponding procedure. The different algorithms and corpus types uniquely characterizing each procedure are for their part represented by distinct colors and letter tags, in that order. Regarding the color coding of algorithms in particular, given the fact that FastText and FastSpell algorithms perform identically for most linguistic categories considered, a different and common color was used for all languages in which performance scores were identical for both algorithms (i.e. all categories but Other languages). Finally, in order to compare overall linguistic performances, harmonic averages of precision and recall scores by language for all procedures implemented for this project are represented in each subplot by horizontal and vertical dashed lines respectively.

\afterpage{
\begin{figure}[ht!]
\centering
\includegraphics[width=.85\textwidth]{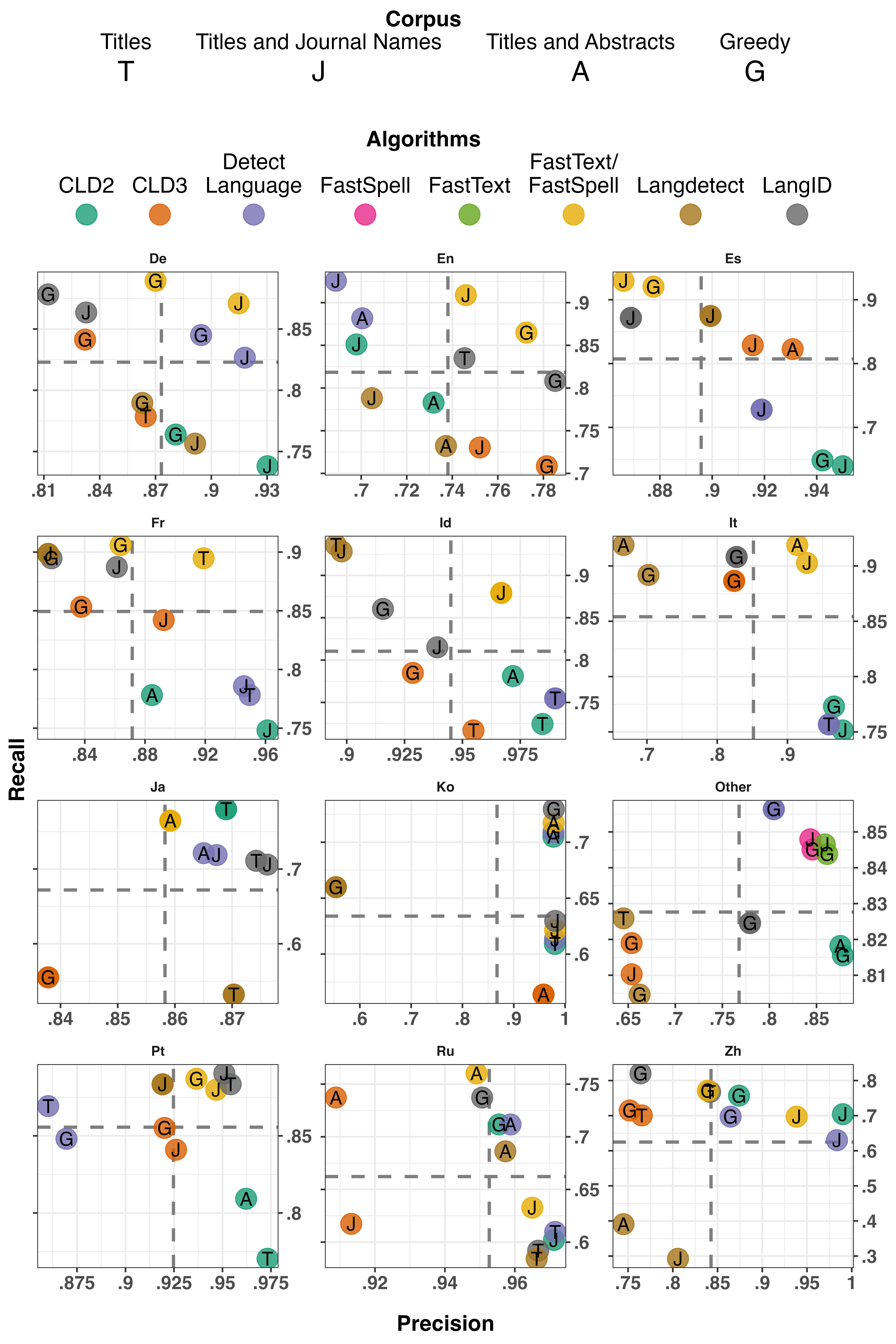}
\caption{Proportion and Recall Scores for all Procedures, grouped by Language and Corpus Type}
\label{fig:perf}
\end{figure}
}

As the different subplots show, procedure performance varies both substantially and differently across the different subplots, with each algorithm or corpus type performing more or less well depending on the specific circumstances of each language subset evaluated. Looking at procedure score averages for each language, precision scores are higher than their recall counterparts except for English and Other articles. Average precision is highest for Russian and Indonesian articles, while the lowest scores belong to articles written in English and Other languages; as for recall, averages are highest for French, Italian, and Portuguese articles, and lowest in the case of Korean, Japanese, Chinese, and Russian articles. Overall, the performances obtained confirm the well-known inverse relationship between precision and recall: use of the two-sided Wetzels Bayes Factor shows that both performance measures are inversely correlated at a rate of $\rho = -.18$, with the observed data supporting the presence of correlation over the null hypothesis at a ratio of 127437 to 1 (Wetzels et al., 2012).

With regard to the different types of corpus implemented, all corpus types have procedures with above-average precision and recall scores in at least one language. While the various filtering procedures described above prevent any in-depth analysis of performance by corpus types, much more can however be said of performances by algorithms and languages. First and foremost, the first quadrant of the plane determined by the two perpendicular lines representing the precision and recall averages include points in all cases, which means that every language includes at least one procedure which is above average in both precision and recall. Moreover, each algorithm has procedures that score above both precision and recall average in at least one language: while FastText and FastSpell have at least one procedure with above-average precision and recall scores in all languages but Spanish and Russian, Spanish is the only language where the CLD3 algorithm scores above average in both precision and recall. In addition, two algorithms (Langdetect for articles in English, Chinese, and Other languages as well as CLD3 for Japanese and Other languages) have procedures whose precision and recall scores are both below language average.

As regards to each measure considered separately, no single algorithm has above-average precision scores in all languages, while FastSpell and FastText are the only algorithms with at least one above-average recall score in each language category. These two algorithms are also those that stand out the most across languages, with procedures performing above precision or recall average in all languages except Spanish and Russian. In the case of French, Indonesian, and Italian articles, FastSpell and FastText are actually the only algorithms whose procedures score above average for both measures. Regarding the Other  category, which is the only one in which the performance of both algorithms differ, FastText performs markedly better in terms of precision, while recall scores of FastSpell are slightly higher than those of FastText; for all other languages, the additional features of Fast Spell have no impact other than increasing the space and time complexity of the procedures involved.

Finally, a note regarding English-written articles, which account for the vast majority of articles indexed in the database. FastText, FastSpell, and to a lower extent LangID are the only algorithms recording above-average scores in both precision and recall; besides these algorithms, DetectLanguage and CLD2 have procedures that score above-average in terms of recall, while CLD3 is the only other algorithm with above-average precision scores. The highest precision score is recorded by the LangID algorithm on the Greedy corpus, while the best recall performance is obtained by applying the DetectLanguage algorithm on Titles \& Journal Names. However, as both procedures have relatively low scores in the case of the other measure, it can be concluded that procedures based on the FastSpell and FastText algorithms (in conjunction with the one consisting in the use of the LangID on the Titles corpus, but to a lesser extent) are those whose overall performance stand out the most for English. Given the hegemony of that language in scientific publishing and its prevalence in the database, we should also expect these procedures to perform relatively well in the large-scale simulation that follows. However, in light of the peculiarities mentioned in the Methods section regarding the precision-recall trade-offs in imbalanced datasets, the linguistic classification performance of English-written articles cannot be assessed alone, independently of those of all other languages present in the database. On the contrary, a linguistic classification of all articles indexed in the OpenAlex database can only be effective insofar at it integrates all languages and weighs its impact on each of them in a way similar in scope and aim to the simulation conducted here and whose results are presented and discussed below.

\subsection{Processing Times}

Figure \ref{fig:speed} shows the normalized processing times for the different algorithms and corpus types; a table showing the absolute processing times recorded for all classification procedures implemented can be found in appendix \ref{appendix:speed}. To improve plot readability, vertical line segments connecting all processing times obtained by each algorithm for the various corpora have been added. Fastest and slowest processing times for each corpus type are shown in seconds at both extremities of the x-axis. Finally, in order to better highlight the differences between both slower and faster algorithms, the x-axis has been rescaled using arcsine transformation.

\afterpage{
\begin{figure}[ht!]
\centering
\includegraphics[width=.85\textwidth]{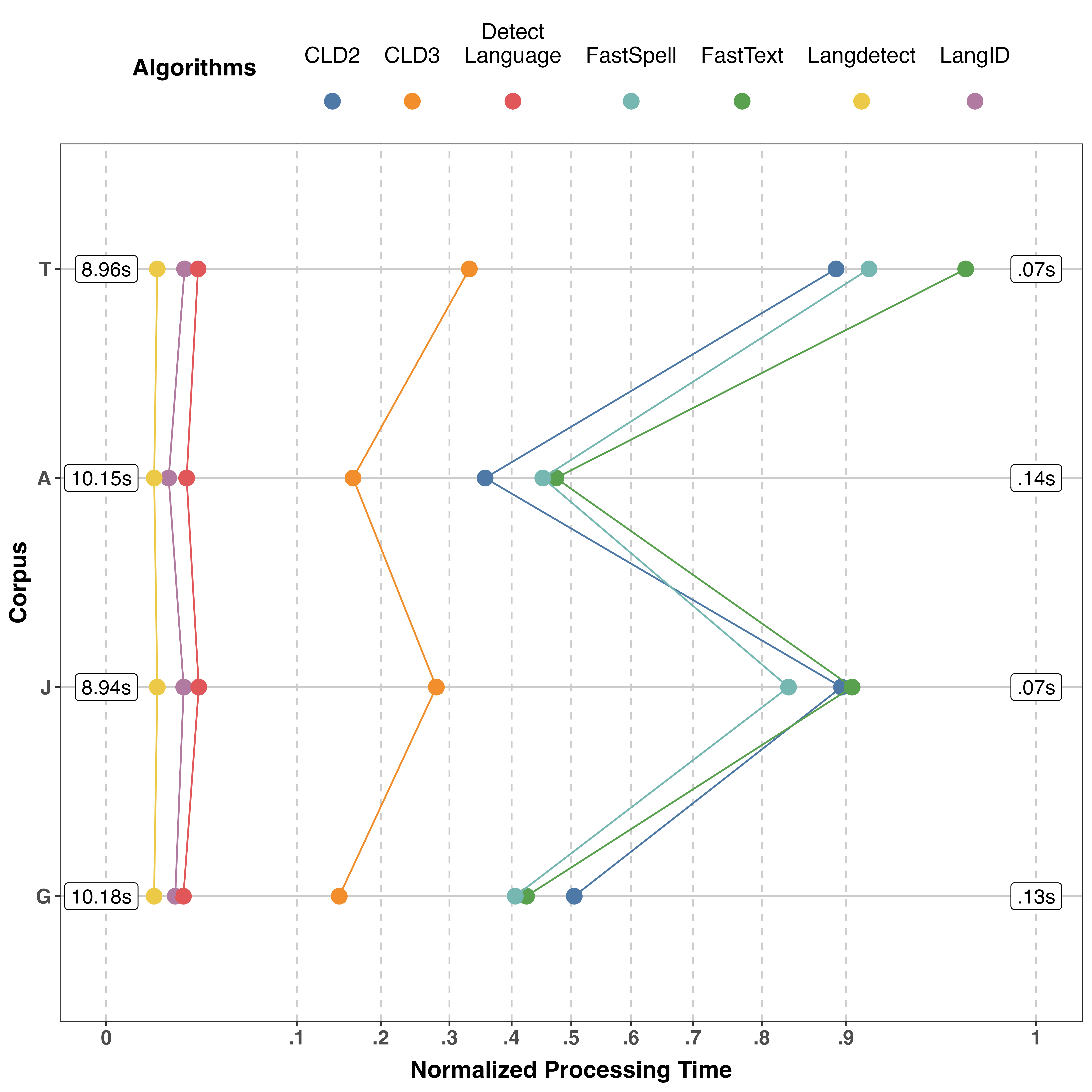}
\caption{Normalized Processing Times by Algorithm and Corpus}
\label{fig:speed}
\end{figure}
}

In terms of processing times, the algorithms implemented and evaluated in this project form three distinct groups whose performances remain consistent across the different corpus types. Langdetect, LangID and DetectLanguage are by far the slowest algorithms in the group, with respective performances of 38.24, 18.61,  and 13.07 seconds for all four corpora assessed. The fastest group includes FastText, FastSpell and CLD2, whose overall performances of 0.44, 0.46, and 0.47 seconds are two orders of magnitude faster than those of the first group, for all corpus types. The third and last group solely consists of CLD3, whose overall performance of 1.29 seconds is an order of magnitude faster and slower than the first and second groups respectively.

In the context of the present project, such differences in performance makes processing speed a crucial dimension of linguistic classification performance. Indeed, while a difference of a few seconds at the sample level may not seem so crucial at first glance, the sheer size of OpenAlex implies that processing speed differences beyond one order of magnitude might likely result in delays ranging from several days to a few weeks. Given the fact that the OpenAlex database is continuously updated with new publication data, such delays can cause serious feasability issues. 

Regarding the performances of FastSpell and FastText in particular, the former is slightly slower than the latter in all cases, which can be expected given that FastSpell adds an upstream spelling checking functionality to FastText. As shown in the previous figure, this enhancement translates into a gain in recall performance for languages included in the Other category. The question remains, however, whether these local gains in recall performance are worth the resulting costs in both precision and speed.

As for DetectLanguage, the fact that the algorithm can only be used through API queries has an undeniable impact on the processing speed of all procedures using this algorithm: even though the algorithm's batch mode helps reduce the number of queries by processing multiple inputs at a time, the fact that its processing speed ultimately depends on factors that have nothing to do with algorithmic performance /textit{stricto sensu}, for example connection speed or server response time, unfortunately makes it an unviable LI strategy for the OpenAlex database.

To a lesser extent, differences in processing times can also be observed between the various corpus types. Across all algorithms, the Titles (T) and Titles \& Abstracts (A) corpora are the corpus types whose processing were the fastest and slowest respectively, with times of $15.98$ and $20.64$ seconds. Average document size can probably and partly explain such differences between corpus types: since titles tend to be substantially shorter than abstracts, with $72.26$ characters on average to $1238.85$ characters for the latter, corpora that include abstracts will take more time to process than those that do not. From a performance evaluation perspective, a situation similar the one observed for FastSpell and FastText thus arises: while abstracts can be very informative from a LI perspective, taking them into account for that purpose is however bound to slow down algorithms; whether this gain in information results in an overall gain of performance will ultimately depend on its impact on precision and recall scores, which once again proves the crucial relevance of corpus processing times to the present LI algorithm assessment.

\subsection{Simulation Results}

Figure \ref{fig:sim} shows the maximum Maximum $F_{\beta\gamma}$ MAP estimate obtained by each algorithm for every combination of precision ($\beta$) and speed ($\gamma$) weights. Each subfigure presents the results obtained for every $\gamma$ value implemented, while $F_{\beta, \gamma}$ scores and $\beta$ weights obtained for any given $\gamma$ value are respectively shown along the x- and y-axes of the corresponding subfigure. For each subfigure, corpus types and algorithms are uniquely and respectively represented using distinct colors and line styles. Finally, subfigures representing weighing regimes that include processing times ($\gamma > 0$) are grouped together horizontally to increase readability.

\afterpage{
\begin{figure}[ht!]
\centering
\includegraphics[width=\textwidth]{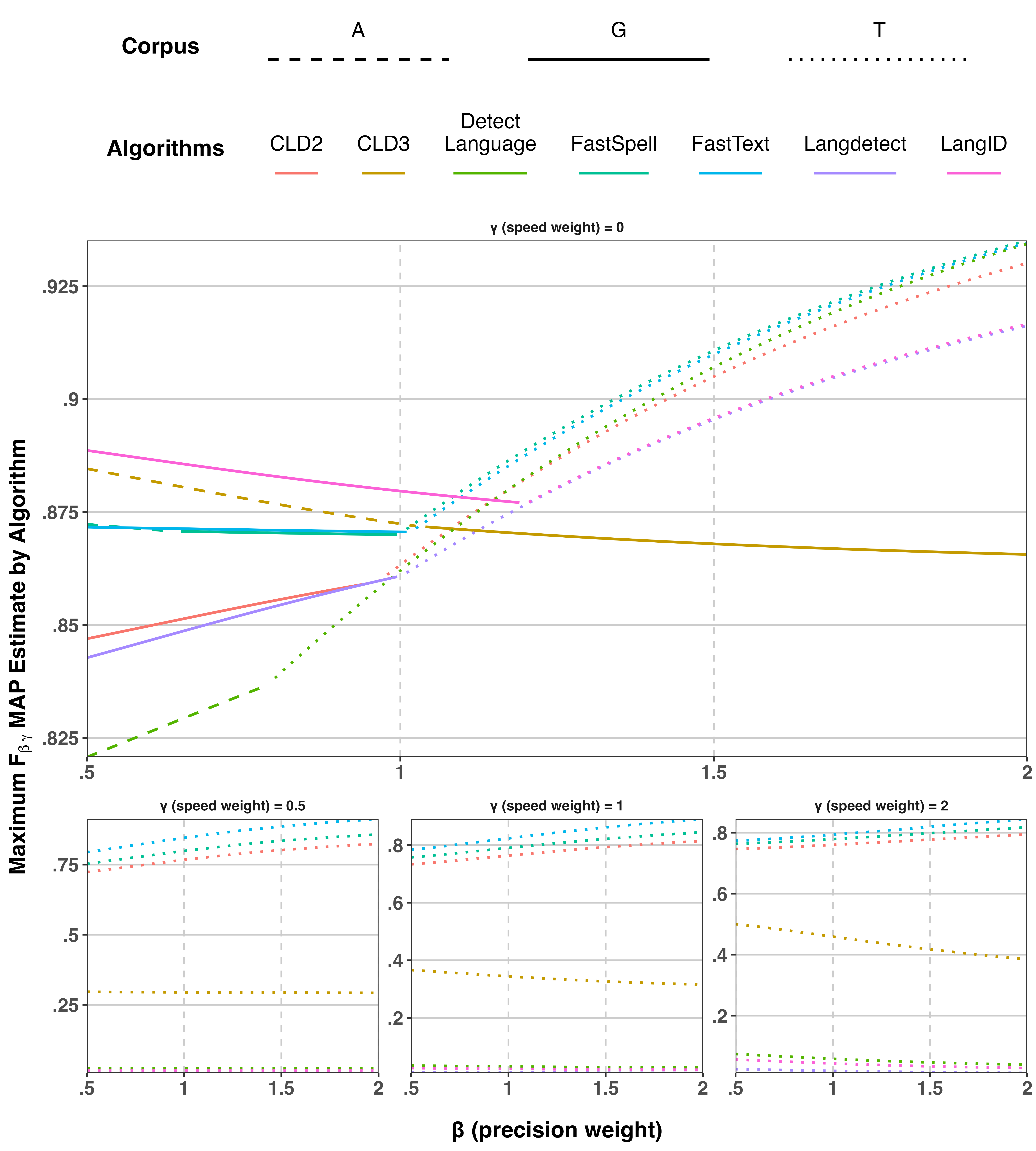}
\caption{Maximum $F_{\beta\gamma}$ MAP Estimate for each algorithm over selected $\beta$ and $\gamma$ values}
\label{fig:sim}
\end{figure}
}

At first glance, what stands out from all the subfigures plotted is the absence of the Titles \& Journal Names (J) corpus. Given the visualization method adopted here, this means that no procedure based on that corpus obtained the highest MAP estimate for any algorithm or weighing regime. This contrasts sharply with the results shown in Fig \ref{fig:perf}, which shows that many procedures based on that corpus perform relatively well for certain languages. In our opinion, such performance shift proves that procedures based on other corpus types have shown greater resilience to uncertainty, as modeled by the simulation process.

Looking then at the topmost subfigure, which shows $F_{\beta, \gamma}$ scores for $\gamma = 0$ (where is not taken into account), best-performing procedures for each algorithm show interesting similarities. First and foremost, performance for all algorithms seem to follow a dual performance regime, with a transition point at around $ \beta \approx 1$, in other words where the weights of precision and recall converge. In the case of DetectLanguage and LangID, this cut-off point occurs a bit earlier ($\beta \approx 0.6$) and later ($\beta \approx 1.2$) along the $\beta$ continuum, but the regime shift is just as sudden and lasting as elsewhere. The first characteristic of that shift is that that all algorithms switch corpus types: whereas the best-performing procedure of CLD3 transitions from the Titles \& Abstracts to the Greedy corpus (i.e. it adds journal names), procedures for all other algorithms become title-based as recall weight increases. The second salient feature of the dual-regime observed at $\gamma = 0$ is that, again with the exception of the exponential decay of CLD3, which seems unaffected by recall weight increase, all best-performing procedures switch to an exponentially-decreasing growth regime as they hit their respective cut-off points and transitions to the titles-based corpus. 

While the performance regimes of best-performing procedures in more precision-oriented settings ($\beta < 1$) are not as striking uniformity as the recall-oriented regimes, some shared properties can nonetheless be observed. In particular, three different groups can be identified based on their corpus types and/or slopes. The first group includes two procedures: a first one that uses LangID on the greedy corpus, and a second one using CLD3 on the Titles \& Abstracts corpus. Showing higher scores than the other procedures, these two procedures however see their performance decrease steadily as greater weight is gradually conferred to recall. The second algorithm group includes FastText and FastSpell, whose above-average performance on the greedy corpus decreases slightly as precision weight gets closer $\beta = 1$. A third group of algorithms that comprises CLD2 and Langdetect is characterized by below-average, yet steadily increasing greedy-corpus-based performances below $\beta < 1$. Finally, the best-performing procedure of DetectLanguage in precision-oriented settings stand out by having the lowest scores and cut-off point of all algorithms, but also the most marked growth of all algorithms, both before and after the cut-off point.

As regards to the other subfigures, i.e. those showing scores that take processing speed into account ($\gamma > 0$), all give a unanimous picture of procedure performance. First, best-performing procedures of each algorithm follow a single, title-based regime for the whole $.5 <= \beta <= 2$ range. But also and perhaps more importantly, CLD2, FastSpell, and especially FastText score markedly higher than all other algorithms, and this outperformance grows substantially as both $\beta$ and $\gamma$ values increase.

In light of these considerations and the simulation results from which they are derived, which classification procedure performs best depends on the importance given to the different evaluation measures implemented. As regards to corpus types, emphasizing precision over recall will favor the more metadata-rich procedures (i.e. the greedy and titles \& abstracts corpora); however, as recall gains in importance and as soon as processing speed is taken into account, the Titles corpus becomes the unanimous choice. As regards to algorithms, while LangID stands out in precision-oriented, speed-overlooking settings, FastText outperforms other algorithms in all other contexts, and the extent of its predominance increases proportionally with both $\beta$ and $\gamma$.

\section{Discussion}

The present project aimed to systematically compare various language classification procedures, procedures combining various Python LI algorithms and metadata-based corpora extracted from manually-annotated articles sampled from the OpenAlex database. Following an analysis of precision and recall performance for each algorithm, corpus, and language as well as of processing speeds recorded for each algorithm and corpus type, overall procedure performance at the database level was simulated using probabilistic confusion matrices for each algorithm, corpus, and language as well as a probabilistic model of relative article language frequencies for the whole OpenAlex database. Results show that procedure performance strongly depends on the importance given to each of the measures implemented: for contexts where precision is preferred, using the LangID algorithm on the greedy corpus gives the best results; however, for all cases where recall is considered at least slightly more important than precision or as soon as processing times are given any kind of consideration, the procedure combining the FastText algorithm and the Titles corpus outperforms all other alternatives. 

In addition to the quality of its overall performance, both LangID and FastText have the advantage of returning as output the probability value of the most probable language. In this sense, a two- or multi-step classification procedure would be quite feasible. For instance, FastText could be run on the titles of all articles indexed in the database, then other procedures that perform better in specific contexts could be run on all cases where the probability/confidence score of the detected language is below a certain threshold. Given the imbalanced nature of the OpenAlex database and the inherent algorithmic tendency to overpredict the majority class, a special effort could then be made to improve prediction for English documents: as mentioned in \citet{cespedes2025evaluating}, English documents are overrepresented in the OpenAlex database at a rate of around 9\%, which negatively affects the representativity of all other languages in the database to a greater or lesser extent, but also reinforces their underrepresentation in the scholarly and knowledge ecosystem at large. In that perspective and given OpenAlex's status as the first truly multilingual database, improving the database's language metadata has social and political implications that go well beyond the functionalities that are normally expected of bibliographic databases.

Results presented and discussed in the previous paragraphs are far from definitive.  For once, they are inextricably tied to the current state-of-the-art in the field of automatic LI. The conclusions drawn are in no way meant to solve the problem of the linguistic classification of OpenAlex articles once and for all, but only to attempt to offer the best possible algorithmic recommendation based on current domain knowledge as well as on the available techniques deemed most suitable for that specific purpose.

Another important point to mention is that since the observations and analyses presented here are the direct result of the methodological choices adopted by the authors, the conclusions derived from the results of this project are valid only within the framework of this precise \textit{modus operandi}. In other words, different choices as regards to the selection of performance measures or the design and parametrization of the simulation process may well have led to different conclusions. For example, the case of the aggregation of performance scores performed here is suggestive in this regard, as increased differences in weighing between precision, recall, and processing speed could lead to different results;  the question remains however as to whether more diversified or extreme weighing regimes than those presented here would be really useful. Nevertheless, a certain epistemic caution is in order here, as the results presented here are far from definitive and should not be considered as such.

As regards to OpenAlex itself, it is also important to note that a significant part of the data contained in the database is provided by the journal publishers themselves based on what they consider relevant bibliometric data. Such decisional idiosyncrasies can have an impact on the performance of monolingual LI algorithms and the language classification they provide, for example if multilingual titles or abstracts are provided for specific journals or articles. This specific scenario in turn points to an important consideration regarding the functionalities of the algorithms considered here, but also those of LI algorithms in general. While current monolingual LI algorithms offer above human-level performance over hundreds of different languages, a current and major challenge in Natural Language Processing consists in designing algorithms that achieve high accuracy when dealing with documents written in multiple languages. Given these considerations, multilingual record entries probably represents the biggest current linguistic classification issue for the OpenAlex database: not only is the quality of linguistic metadata inevitably affected by this problem, but its extent is hard to assess, given the difficulty of identifying which attributes are multilingual and as such likely to cause misclassifications.

Despite these limitations, the results presented here show that, with the right LI procedure, a high-quality linguistic classification of all articles indexed in OpenAlex is nevertheless possible, classification which would not only reinforce the latter's status as the first truly multilingual, large-scale bibliometric database, but also and more generally foster cross-linguistic bibliometric assessments and analyses of unmatched accuracy and comprehensiveness.

Beyond language metadata considerations, we believe the current research project, both in its aims and results, may provide fruitful insights as regards to one one of the biggests problems currently plaguing NLP research, but also science in general, namely the research-practice gap. Summarized in a few words, this gaps refers to the persistent disconnect between what scientific research demonstrates to be effective and what practitioners implement in real-world contexts  \citep{haney2024bridging}. Most often, this gap is defined and manifests itself in temporal terms, more precisely as the delay between the publication of research and its integration into professional practice. Studies in healthcare, education, psychology, organizational science, criminology, and social work have reported multi-year lags in putting evidence into use \citep{greenhalgh2016research, lau2016achieving,ebsco2023, elliott2000bridging}. But the gap between research and practice can also further materialize in the partial, inconsistent, unsustained, unscaled or wrongful application of research results. In the worst case, research remains completely dormant, sometimes due to ignorance, but at other times in full awareness, when factors such as lack of budget, knowledge, trust, training, misaligned incentives or simply inertia or resistance to change are at play \citep{rynes2001research, greenhalgh2016research}. More than a mere communication failure, the research-practice gap is a structural divide between two distinct communities with different goals, values, languages, incentives and constraints \citep{Caplan1979_two_communities}. Across domains.and areas of investigations, any such discrepancy has the potential to create a significant divide between research and practice communities or to worsen an already existing one. Ironically, even implementation science itself is facing implementation issues \citep{westerlund2019researchpractice}.

As regards to the fields of Artificial Intelligence and Machine Learning, multiple large practitioner surveys by Kaggle, Rexer Analytics, and JetBrains report prevalent, widespread, and enduring usage of regression methods (linear, logistic or generalized linear models) among data scientists in the field \citep{rexer2023, Rexer2020_DataScienceSurvey, Kaggle2020_StateOfML, jetbrains2023}. Thus, while academic ML research produces ever more sophisticated algorithms (deep nets, transformers, etc.), industry practitioners frequently default to simpler, established, reliable, and interpretable methods that favor robustness, simplicity, transparency, explainability, and lower deployment costs.

This structural divide is especially salient in the field of Natural Language Processing. Despite the growing needs for language technology across domains, NLP systems that perform well in research settings often fail to fully translate into practical, sustainable, and scalable deployments \citep{gessler-etal-2025-understanding}. Perhaps here more than elsewhere, this deadlock persists because NLP research seeks to optimize conditions that rarely resemble those encountered at the deployment stage, while practical environments are met with challenges and needs which the research ecosystem and its reward system seldom prioritizes nor even considers.

More than a simple problem of communication, dissemination, or access to knowledge, such misalignment is on the contrary present at all stages of the software development lifecycle, from the earliest stages of a project to its deployment and long-term maintenance. At the requirements stage, research typically defines abstract tasks and goals based on gaps or current trends in their field of expertise; at the other end,  healthcare and government adoption studies show that practitioners must satisfy organizational constraints, stakeholder needs, regulatory requirements, and domain-specific risks which are most often completely unrelated with the current state of literature in the relevant research areas \citep{fu2023recommended, jiang2023adoption}. This divergence widens at the data collection and preprocessing: where research builds on static, well-curated benchmarks, practical NLP work must handle messy, heterogeneous, multilingual, and continuously shifting data environments. This “data readiness” gap—long noted by practitioners—means that much of the effort in real-world NLP is spent on data acquisition, cleaning, annotation, documentation, and integration, tasks that the research incentive structure largely overlooks \citep{olsson-sahlgren-2021-datareadiness, joshi2020}. By the time systems reach the modeling and evaluation phase, the mismatch between research and practice is amplified by evaluation regimes that privilege confusion matrix-based performance metrics over robustness, safety, interpretability, and domain fidelity \citep{ribeiro2020checklist, rudin2019stop, jia2017adversarial, bender2021parrots}. As with other research areas, reproducibility considerations are also often overlooked at the evaluation stage: in a compilation of repeatability (reproduction under the same conditions) assessment studies of NLP models , \citet{belz2021systematic} found out that only 14.03\% of all 513 score reproduction attempts were successful, and when reproduction failed, reproduced results were most of the time worse (59.2\%) than in the original study.

The various activities involved at the deployment and maintenance phases also contribute to widen the gap. Prototyping rarely accounts for the engineering and infrastructural issues that might arise when models meet production constraints, such as monitoring, logging, versioning, rollback or performance optimization. Especially for large neural models, other practical considerations such as spatial and temporal complexity, latency, scalability, energy consumption, and security risks can create additional and maybe insurmountable barriers \citep{strubell-etal-2019-energy}. In addition, NLP models and ML in general are far more fragile and difficult to maintain than traditional software, as their unstable data dependencies, brittle pipelines, and tightly coupled components make even small changes risky and costly \citep{sculley2015hidden, breck2017mltestscore}. Finally, once NLP systems reach users, new divergences emerge: while research often assumes that the model's output is the final product, practitioners must design workflows, interfaces, tests, and feedback mechanisms that meet real human needs and allow for safe and iterative improvement \citep{amershi2019guidelines}.

Viewed from a software development perspective, the research–practice gap in NLP thus emerges not from a single failure point but from a lifecycle-long accumulation of mismatches in goals, assumptions, incentives, and constraints. Regardless of model performance considerations, each new project that follows this dynamic simply adds an interaction to this vicious circle and widens the gap between researchers and practitioners. To bridge this divide, NLP research has to strive for more than the usual productivity, originality, and accuracy imperatives; as this project tried to show and achieve, it must also and from the outset be geared towards real-world use. This can only be achieved by engaging early and continuously with end-users, organizations, and policy systems throughout the development process and through initiatives like code sharing, co-design and deployment studies \citep{olsson-sahlgren-2021-datareadiness, lin2022truthfulqa, pillai2023toward, gessler-etal-2025-understanding}. Participatory and community-based approaches, which are massively underrepresented in mainstream NLP research \citep{pillai2023toward}, can also be very effective in ensuring that research initiatives are both sustainable and useful to the users they target. Contrary to what current incentive structures might suggest, the publication of research results is not an end in itself, but rather the medium through which researchers and practitioners can collaborate to make research truly impactful and innovative. This project, by attempting to provide a comprehensive assessment of automatic language detection algorithms based on the current structure and needs of OpenAlex, aims to contribute to this realignment of NLP research with the objectives and needs of practitioners, likely to reshape research agendas and priorities towards more concretely transformative and truly meliorative outputs.

\bibliography{manuscript.bib}

\backmatter

\bmhead{Acknowledgements}

We would like to thank all of our colleagues who contributed to the annotation of the article sample as part of the article by \citet{cespedes2025evaluating}. This article stems from the reflections and
experiences they shared throughout the annotation process. We also thank Juan Pablo Alperin for his suggestion regarding the use of multi-step language identification procedures.

\section*{Declarations}

\begin{itemize}
\item Lucia Céspedes, Diego Kozlowski, and Vincent Larivière acknowledge funding from the Social Science and Humanities Research Council of Canada Pan-Canadian Knowledge Access Initiative Grant (Grant Number 1007-2023-0001), and the Fonds de recherche du Québec—Société et Culture through the Programme d'appui aux Chaires UNESCO (Grant Number 338828)
\item The authors declare no conflicts of interest.
\item The data generated and analyzed in this study can either be found in the appendices or in the article by \citet{cespedes2025evaluating}
\item The code used in this study is available from the corresponding author upon request.
\item \textbf{CRediT authorship contribution statement.} \textbf{Maxime Holmberg Sainte-Marie}: Conceptualization, Data Curation, Formal Analysis, Investigation, Method-
ology, Project Administration, Software, Validation, Visualization, Writing - Original Draft; \textbf{Diego Kozlowski}: Conceptualization, Formal Analysis, Methodology; \textbf{Lucía Céspedes:}
Conceptualization, Project Administration.
Validation, Visualization; \textbf{Vincent Larivière:} Conceptualization, Project Administration, Funding Acquisition, Re-
sources, Supervision.
\end{itemize}
\clearpage
\begin{appendices}

\section{OpenAlex linguistic distribution, grouped by metadata availability}
\label{appendix:distribution}

The following table shows article frequency by metadata availability for the different language categories used in this project, as of December 25th, 2024.

\begin{table}[ht!]
\centering
\begin{tabular}{r|cccc}
\hline
\textbf{Language} & \textbf{Title} & \makecell{\textbf{Title \&} \\\textbf{Abstract}} & \makecell{\textbf{Title \&}\\\textbf{Journal Name}} & \makecell{\textbf{Title, Abstract \&}\\\textbf{Journal Name}}\\
\hline
de & 615 864 & 208 947 & 1 981 603 & 1 154 153\\
en & 17 915 165 & 16 847 352 & 43 085 593 & 69 649 893\\
es & 1 918 418 & 689 230 & 2 207 249 & 1 948 762\\
fr & 1 010 112 & 501 142 & 2 096 853 & 1 209 886\\
id & 308 631 & 402 811 & 113 823 & 1 332 430\\
it & 762 538 & 67 312 & 514 968 & 107 421\\
ja & 952 769 & 6 185 & 6 834 274 & 214 832\\
ko & 600 926 & 87 884 & 2 446 512 & 1 140 439\\
pt & 657 215 & 409 953 & 476 663 & 1 588 976\\
ru & 360 858 & 79 011 & 1 175 999 & 597 653\\
zh & 100 936 & 188 780 & 912 446 & 3 472 426\\
other & 2 050 859 & 818 517 & 2 922 167 & 1 724 352\\
unknown & 930 370 & 14 522 & 3 863 730 & 246 058\\
\hline
\end{tabular}
\end{table}

\section{Processing times}
\label{appendix:speed}

The following table shows processing times for each algorithm and corpus, in elapsed seconds.

\begin{table}[ht!]
\centering
\begin{tabular}{r|ccccccc}
\rot{45}{\textbf{Corpus}} & \rot{45}{\textbf{CLD2}} & \rot{45}{\textbf{CLD3}} & \rot{45}{\makecell{\textbf{Detect}\\\textbf{
Language}}} & \rot{45}{\textbf{FastSpell}} & \rot{45}{\textbf{FastText}} & \rot{45}{\textbf{LangID}} & \rot{45}{\textbf{Langdetect}}\\
\hline
T & 0.07 & 0.20 & 2.78 & 0.07 & 0.07 & 3.82 & 8.96\\
A & 0.19 & 0.40 & 3.62 & 0.15 & 0.14 & 5.99 & 10.15\\
J & 0.07 & 0.24 & 2.74 & 0.08 & 0.07 & 3.90 & 8.94\\
G & 0.13 & 0.45 & 3.93 & 0.16 & 0.16 & 4.91 & 10.18\\
\hline
\end{tabular}
\end{table}

\clearpage

\section{Performance by Language}
\label{appendix:perf}

The following tables show precision and recall scores for the different languages, grouped by corpus and algorithm.

\subsection{Chinese}

\begin{table}[ht!]
\centering
\begin{tabular}[t]{c|ccccccc}
\rot{45}{\textbf{Corpus}} & \rot{45}{\textbf{CLD2}} & \rot{45}{\textbf{CLD3}} & \rot{45}{\makecell{\textbf{Detect}\\\textbf{
Language}}} & \rot{45}{\textbf{FastSpell}} & \rot{45}{\textbf{FastText}} & \rot{45}{\textbf{LangID}} & \rot{45}{\textbf{Langdetect}}\\
\hline\hline
\multicolumn{8}{c}{\textbf{Precision}}\\
\hline

T & 0.98 & 0.77 & 0.98 & 0.93 & 0.93 & 0.84 & 0.78\\
A & 0.87 & 0.75 & 0.86 & 0.84 & 0.84 & 0.77 & 0.74\\
J & 0.99 & 0.76 & 0.98 & 0.94 & 0.94 & 0.81 & 0.81\\
G & 0.87 & 0.75 & 0.86 & 0.84 & 0.84 & 0.76 & 0.76\\
\hline\hline
\multicolumn{8}{c}{\textbf{Recall}}\\
\hline

G & 0.76 & 0.71 & 0.70 & 0.77 & 0.77 & 0.82 & 0.37\\

T & 0.69 & 0.70 & 0.62 & 0.69 & 0.69 & 0.77 & 0.34\\

A & 0.75 & 0.70 & 0.70 & 0.77 & 0.77 & 0.81 & 0.39\\

J & 0.70 & 0.70 & 0.63 & 0.70 & 0.70 & 0.78 & 0.29\\

\hline
\end{tabular}
\end{table}

\subsection{English}

\begin{table}[ht!]
\centering
\begin{tabular}[t]{c|ccccccc}
\rot{45}{\textbf{Corpus}} & \rot{45}{\textbf{CLD2}} & \rot{45}{\textbf{CLD3}} & \rot{45}{\makecell{\textbf{Detect}\\\textbf{
Language}}} & \rot{45}{\textbf{FastSpell}} & \rot{45}{\textbf{FastText}} & \rot{45}{\textbf{LangID}} & \rot{45}{\textbf{Langdetect}}\\
\hline\hline
\multicolumn{8}{c}{\textbf{Precision}}\\
\hline

G & 0.72 & 0.78 & 0.70 & 0.77 & 0.77 & 0.79 & 0.72\\
T & 0.71 & 0.74 & 0.69 & 0.74 & 0.74 & 0.75 & 0.72\\
A & 0.73 & 0.78 & 0.70 & 0.77 & 0.77 & 0.78 & 0.74\\
J & 0.70 & 0.75 & 0.69 & 0.75 & 0.75 & 0.76 & 0.70\\
\hline\hline
\multicolumn{8}{c}{\textbf{Recall}}\\
\hline
G & 0.81 & 0.71 & 0.89 & 0.86 & 0.86 & 0.81 & 0.76\\
T & 0.83 & 0.70 & 0.92 & 0.90 & 0.90 & 0.83 & 0.75\\
A & 0.78 & 0.69 & 0.88 & 0.86 & 0.86 & 0.80 & 0.73\\
J & 0.85 & 0.73 & 0.93 & 0.91 & 0.91 & 0.83 & 0.79\\
\hline
\end{tabular}
\end{table}
\clearpage

\subsection{French}

\begin{table}[ht!]
\centering
\begin{tabular}[t]{c|ccccccc}
\rot{45}{\textbf{Corpus}} & \rot{45}{\textbf{CLD2}} & \rot{45}{\textbf{CLD3}} & \rot{45}{\makecell{\textbf{Detect}\\\textbf{
Language}}} & \rot{45}{\textbf{FastSpell}} & \rot{45}{\textbf{FastText}} & \rot{45}{\textbf{LangID}} & \rot{45}{\textbf{Langdetect}}\\
\hline\hline
\multicolumn{8}{c}{\textbf{Precision}}\\
\hline
G & 0.90 & 0.84 & 0.92 & 0.86 & 0.86 & 0.82 & 0.76\\
T & 0.96 & 0.89 & 0.95 & 0.92 & 0.92 & 0.86 & 0.81\\
A & 0.88 & 0.84 & 0.92 & 0.86 & 0.86 & 0.81 & 0.76\\
J & 0.96 & 0.89 & 0.95 & 0.92 & 0.92 & 0.86 & 0.82\\
\hline\hline
\multicolumn{8}{c}{\textbf{Recall}}\\
\hline
G & 0.77 & 0.85 & 0.78 & 0.91 & 0.91 & 0.89 & 0.89\\
T & 0.76 & 0.84 & 0.78 & 0.89 & 0.89 & 0.89 & 0.89\\
A & 0.78 & 0.85 & 0.77 & 0.90 & 0.90 & 0.89 & 0.89\\
J & 0.75 & 0.84 & 0.79 & 0.90 & 0.90 & 0.89 & 0.90\\
\hline
\end{tabular}
\end{table}

\subsection{German}

\begin{table}[ht!]
\centering
\begin{tabular}[t]{c|ccccccc}
\rot{45}{\textbf{Corpus}} & \rot{45}{\textbf{CLD2}} & \rot{45}{\textbf{CLD3}} & \rot{45}{\makecell{\textbf{Detect}\\\textbf{
Language}}} & \rot{45}{\textbf{FastSpell}} & \rot{45}{\textbf{FastText}} & \rot{45}{\textbf{LangID}} & \rot{45}{\textbf{Langdetect}}\\
\hline
\multicolumn{8}{c}{\textbf{Precision}}\\
\hline
G & 0.88 & 0.83 & 0.89 & 0.87 & 0.87 & 0.81 & 0.86\\
T & 0.93 & 0.86 & 0.91 & 0.91 & 0.91 & 0.83 & 0.86\\
A & 0.88 & 0.83 & 0.89 & 0.87 & 0.87 & 0.80 & 0.85\\
J & 0.93 & 0.86 & 0.92 & 0.91 & 0.91 & 0.83 & 0.89\\
\hline\hline
\multicolumn{8}{c}{\textbf{Recall}}\\
\hline
G & 0.76 & 0.84 & 0.85 & 0.89 & 0.89 & 0.88 & 0.79\\
T & 0.73 & 0.78 & 0.82 & 0.86 & 0.86 & 0.86 & 0.75\\
A & 0.76 & 0.82 & 0.84 & 0.89 & 0.89 & 0.87 & 0.79\\
J & 0.74 & 0.81 & 0.83 & 0.87 & 0.87 & 0.86 & 0.76\\
\hline
\end{tabular}
\end{table}

\clearpage

\subsection{Indonesian}

\begin{table}[ht!]
\centering
\begin{tabular}[t]{c|ccccccc}
\rot{45}{\textbf{Corpus}} & \rot{45}{\textbf{CLD2}} & \rot{45}{\textbf{CLD3}} & \rot{45}{\makecell{\textbf{Detect}\\\textbf{
Language}}} & \rot{45}{\textbf{FastSpell}} & \rot{45}{\textbf{FastText}} & \rot{45}{\textbf{LangID}} & \rot{45}{\textbf{Langdetect}}\\
\hline\hline
\multicolumn{8}{c}{\textbf{Precision}}\\
\hline
G & 0.97 & 0.93 & 0.98 & 0.95 & 0.95 & 0.92 & 0.89\\
T & 0.98 & 0.95 & 0.99 & 0.96 & 0.96 & 0.93 & 0.90\\
A & 0.97 & 0.93 & 0.98 & 0.94 & 0.94 & 0.91 & 0.88\\
J & 0.98 & 0.93 & 0.99 & 0.97 & 0.97 & 0.94 & 0.90\\
\hline\hline
\multicolumn{8}{c}{\textbf{Recall}}\\
\hline
G & 0.78 & 0.78 & 0.74 & 0.87 & 0.87 & 0.86 & 0.90\\
T & 0.72 & 0.72 & 0.75 & 0.87 & 0.87 & 0.78 & 0.94\\
A & 0.78 & 0.78 & 0.72 & 0.87 & 0.87 & 0.85 & 0.90\\
J & 0.69 & 0.71 & 0.75 & 0.88 & 0.88 & 0.82 & 0.93\\
\hline
\end{tabular}
\end{table}

\subsection{Italian}

\begin{table}[ht!]
\centering
\begin{tabular}[t]{c|ccccccc}
\rot{45}{\textbf{Corpus}} & \rot{45}{\textbf{CLD2}} & \rot{45}{\textbf{CLD3}} & \rot{45}{\makecell{\textbf{Detect}\\\textbf{
Language}}} & \rot{45}{\textbf{FastSpell}} & \rot{45}{\textbf{FastText}} & \rot{45}{\textbf{LangID}} & \rot{45}{\textbf{Langdetect}}\\
\hline\hline
\multicolumn{8}{c}{\textbf{Precision}}\\
\hline
G & 0.97 & 0.82 & 0.96 & 0.92 & 0.92 & 0.83 & 0.70\\
T & 0.97 & 0.76 & 0.96 & 0.92 & 0.92 & 0.81 & 0.64\\
A & 0.96 & 0.79 & 0.96 & 0.91 & 0.91 & 0.82 & 0.67\\
J & 0.98 & 0.80 & 0.95 & 0.93 & 0.93 & 0.82 & 0.67\\
\hline\hline
\multicolumn{8}{c}{\textbf{Recall}}\\
\hline
G & 0.77 & 0.89 & 0.75 & 0.91 & 0.91 & 0.91 & 0.89\\
T & 0.75 & 0.87 & 0.76 & 0.91 & 0.91 & 0.90 & 0.90\\
A & 0.77 & 0.89 & 0.76 & 0.92 & 0.92 & 0.91 & 0.92\\
J & 0.75 & 0.87 & 0.75 & 0.90 & 0.90 & 0.90 & 0.90\\
\hline
\end{tabular}
\end{table}
\clearpage

\subsection{Japanese}

\begin{table}[ht!]
\centering
\begin{tabular}[t]{c|ccccccc}
\rot{45}{\textbf{Corpus}} & \rot{45}{\textbf{CLD2}} & \rot{45}{\textbf{CLD3}} & \rot{45}{\makecell{\textbf{Detect}\\\textbf{
Language}}} & \rot{45}{\textbf{FastSpell}} & \rot{45}{\textbf{FastText}} & \rot{45}{\textbf{LangID}} & \rot{45}{\textbf{Langdetect}}\\
\hline\hline
\multicolumn{8}{c}{\textbf{Precision}}\\
\hline
G & 0.87 & 0.84 & 0.87 & 0.86 & 0.86 & 0.88 & 0.86\\
T & 0.87 & 0.83 & 0.86 & 0.84 & 0.84 & 0.87 & 0.87\\
A & 0.87 & 0.83 & 0.87 & 0.86 & 0.86 & 0.87 & 0.86\\
J & 0.87 & 0.83 & 0.87 & 0.84 & 0.84 & 0.88 & 0.86\\
\hline\hline
\multicolumn{8}{c}{\textbf{Recall}}\\
\hline
G & 0.78 & 0.55 & 0.72 & 0.76 & 0.76 & 0.71 & 0.53\\
T & 0.78 & 0.55 & 0.72 & 0.76 & 0.76 & 0.71 & 0.53\\
A & 0.78 & 0.55 & 0.72 & 0.76 & 0.76 & 0.71 & 0.52\\
J & 0.78 & 0.55 & 0.72 & 0.76 & 0.76 & 0.71 & 0.52\\
\hline
\end{tabular}
\end{table}

\subsection{Korean}

\begin{table}[ht!]
\centering
\begin{tabular}[t]{c|ccccccc}
\rot{45}{\textbf{Corpus}} & \rot{45}{\textbf{CLD2}} & \rot{45}{\textbf{CLD3}} & \rot{45}{\makecell{\textbf{Detect}\\\textbf{
Language}}} & \rot{45}{\textbf{FastSpell}} & \rot{45}{\textbf{FastText}} & \rot{45}{\textbf{LangID}} & \rot{45}{\textbf{Langdetect}}\\
\hline\hline
\multicolumn{8}{c}{\textbf{Precision}}\\
\hline

G & 0.98 & 0.91 & 0.98 & 0.98 & 0.98 & 0.98 & 0.55\\

T & 0.98 & 0.94 & 0.98 & 0.98 & 0.98 & 0.98 & 0.52\\

A & 0.98 & 0.96 & 0.98 & 0.98 & 0.98 & 0.98 & 0.54\\

J & 0.98 & 0.90 & 0.98 & 0.98 & 0.98 & 0.98 & 0.51\\

\hline\hline
\multicolumn{8}{c}{\textbf{Recall}}\\
\hline

G & 0.70 & 0.56 & 0.71 & 0.72 & 0.72 & 0.73 & 0.66\\
T & 0.61 & 0.53 & 0.60 & 0.62 & 0.62 & 0.63 & 0.56\\
A & 0.70 & 0.56 & 0.70 & 0.72 & 0.72 & 0.73 & 0.66\\
J & 0.61 & 0.52 & 0.61 & 0.62 & 0.62 & 0.63 & 0.56\\
\hline
TJeTAeT & 0.61 & 0.53 & 0.61 & 0.62 & 0.62 & 0.63 & 0.57\\
\hline
\end{tabular}
\end{table}

\clearpage

\subsection{Portuguese}

\begin{table}[ht!]
\centering
\begin{tabular}[t]{c|ccccccc}
\rot{45}{\textbf{Corpus}} & \rot{45}{\textbf{CLD2}} & \rot{45}{\textbf{CLD3}} & \rot{45}{\makecell{\textbf{Detect}\\\textbf{
Language}}} & \rot{45}{\textbf{FastSpell}} & \rot{45}{\textbf{FastText}} & \rot{45}{\textbf{LangID}} & \rot{45}{\textbf{Langdetect}}\\
\hline\hline
\multicolumn{8}{c}{\textbf{Precision}}\\
\hline

G & 0.96 & 0.92 & 0.87 & 0.94 & 0.94 & 0.94 & 0.92\\

T & 0.97 & 0.92 & 0.86 & 0.94 & 0.94 & 0.95 & 0.90\\

A & 0.96 & 0.92 & 0.85 & 0.93 & 0.93 & 0.95 & 0.89\\

J & 0.97 & 0.93 & 0.87 & 0.95 & 0.95 & 0.95 & 0.92\\

\hline\hline
\multicolumn{8}{c}{\textbf{Recall}}\\
\hline

G & 0.80 & 0.86 & 0.85 & 0.89 & 0.89 & 0.88 & 0.86\\
T & 0.77 & 0.83 & 0.87 & 0.87 & 0.87 & 0.88 & 0.87\\
A & 0.81 & 0.85 & 0.85 & 0.88 & 0.88 & 0.88 & 0.86\\
J & 0.77 & 0.84 & 0.86 & 0.88 & 0.88 & 0.89 & 0.88\\

\hline
\end{tabular}
\end{table}

\subsection{Russian}

\begin{table}[ht!]
\centering
\begin{tabular}[t]{c|ccccccc}
\rot{45}{\textbf{Corpus}} & \rot{45}{\textbf{CLD2}} & \rot{45}{\textbf{CLD3}} & \rot{45}{\makecell{\textbf{Detect}\\\textbf{
Language}}} & \rot{45}{\textbf{FastSpell}} & \rot{45}{\textbf{FastText}} & \rot{45}{\textbf{LangID}} & \rot{45}{\textbf{Langdetect}}\\
\hline\hline
\multicolumn{8}{c}{\textbf{Precision}}\\
\hline
G & 0.96 & 0.91 & 0.96 & 0.95 & 0.95 & 0.95 & 0.95\\
T & 0.97 & 0.91 & 0.97 & 0.96 & 0.96 & 0.97 & 0.97\\
A & 0.96 & 0.91 & 0.96 & 0.95 & 0.95 & 0.95 & 0.96\\
J & 0.97 & 0.91 & 0.97 & 0.96 & 0.96 & 0.96 & 0.96\\
\hline\hline
\multicolumn{8}{c}{\textbf{Recall}}\\
\hline

G & 0.71 & 0.73 & 0.70 & 0.75 & 0.75 & 0.74 & 0.68\\
T & 0.58 & 0.61 & 0.61 & 0.63 & 0.63 & 0.59 & 0.58\\
A & 0.71 & 0.74 & 0.71 & 0.76 & 0.76 & 0.73 & 0.69\\
J & 0.60 & 0.62 & 0.61 & 0.63 & 0.63 & 0.63 & 0.58\\
\hline
\end{tabular}
\end{table}

\clearpage

\subsection{Spanish}

\begin{table}[ht!]
\centering
\begin{tabular}[t]{c|ccccccc}
\rot{45}{\textbf{Corpus}} & \rot{45}{\textbf{CLD2}} & \rot{45}{\textbf{CLD3}} & \rot{45}{\makecell{\textbf{Detect}\\\textbf{
Language}}} & \rot{45}{\textbf{FastSpell}} & \rot{45}{\textbf{FastText}} & \rot{45}{\textbf{LangID}} & \rot{45}{\textbf{Langdetect}}\\
\hline\hline
\multicolumn{8}{c}{\textbf{Precision}}\\
\hline
G & 0.94 & 0.92 & 0.91 & 0.88 & 0.88 & 0.87 & 0.89\\
T & 0.94 & 0.93 & 0.92 & 0.86 & 0.86 & 0.86 & 0.89\\
A & 0.94 & 0.93 & 0.91 & 0.87 & 0.87 & 0.86 & 0.90\\
J & 0.95 & 0.92 & 0.92 & 0.87 & 0.87 & 0.87 & 0.90\\
\hline\hline
\multicolumn{8}{c}{\textbf{Recall}}\\
\hline
G & 0.65 & 0.83 & 0.70 & 0.92 & 0.92 & 0.86 & 0.87\\
T & 0.62 & 0.82 & 0.73 & 0.93 & 0.93 & 0.86 & 0.86\\
A & 0.64 & 0.82 & 0.70 & 0.92 & 0.92 & 0.86 & 0.83\\
J & 0.64 & 0.83 & 0.73 & 0.93 & 0.93 & 0.87 & 0.87\\
\hline
\end{tabular}
\end{table}

\subsection{Other Languages}

\begin{table}[ht!]
\centering
\begin{tabular}[t]{c|ccccccc}
\rot{45}{\textbf{Corpus}} & \rot{45}{\textbf{CLD2}} & \rot{45}{\textbf{CLD3}} & \rot{45}{\makecell{\textbf{Detect}\\\textbf{
Language}}} & \rot{45}{\textbf{FastSpell}} & \rot{45}{\textbf{FastText}} & \rot{45}{\textbf{LangID}} & \rot{45}{\textbf{Langdetect}}\\
\hline\hline
\multicolumn{8}{c}{\textbf{Precision}}\\
\hline
G & 0.88 & 0.65 & 0.80 & 0.85 & 0.86 & 0.78 & 0.66\\
T & 0.87 & 0.65 & 0.79 & 0.84 & 0.85 & 0.75 & 0.65\\
A & 0.88 & 0.65 & 0.80 & 0.84 & 0.86 & 0.77 & 0.65\\
J & 0.87 & 0.65 & 0.80 & 0.84 & 0.86 & 0.77 & 0.65\\
\hline\hline
\multicolumn{8}{c}{\textbf{Recall}}\\
\hline
G & 0.82 & 0.82 & 0.86 & 0.85 & 0.84 & 0.82 & 0.80\\
T & 0.81 & 0.81 & 0.86 & 0.84 & 0.84 & 0.81 & 0.83\\
A & 0.82 & 0.82 & 0.85 & 0.84 & 0.84 & 0.82 & 0.82\\
J & 0.81 & 0.81 & 0.85 & 0.85 & 0.85 & 0.82 & 0.80\\
\hline
\end{tabular}
\end{table}

\end{appendices}
\end{document}